\documentclass[10pt]{article}

\usepackage{cite}
\usepackage{amsmath,amssymb,amsfonts}
\usepackage{algorithmic}
\usepackage{hyperref}
\usepackage{graphicx}
\usepackage{textcomp}
\usepackage[frozencache]{minted}
\usepackage[T1]{fontenc}
\usepackage[scaled=0.85]{beramono}
\usepackage{xcolor}
\usepackage{authblk}

\def\BibTeX{{\rm B\kern-.05em{\sc i\kern-.025em b}\kern-.08em
    T\kern-.1667em\lower.7ex\hbox{E}\kern-.125emX}}
\begin{document}

\title{A flexible and fast PyTorch toolkit for simulating training and

  inference on analog crossbar arrays\footnote{submitted to AICAS 2021}}
\author[1]{Malte J. Rasch}
\author[1]{Diego Moreda}
\author[1]{Tayfun Gokmen}
\author[1]{Manuel Le Gallo}
\author[1]{Fabio Carta}
\author[1]{Cindy  Goldberg}
\author[1]{Kaoutar El Maghraoui}
\author[1]{Abu Sebastian}
\author[1]{Vijay Narayanan}

\affil[1]{IBM Research}

\maketitle

\newcommand{\figref}[1]{Fig.~\ref{fig:#1}}
\newcommand{\tabref}[1]{Tab.~\ref{tab:#1}}

\begin{abstract}
  We introduce the \textsc{IBM Analog Hardware Acceleration Kit}, a new and first of a kind open source toolkit to simulate analog crossbar arrays in a convenient fashion from within
  \textsc{PyTorch} (freely available at \url{https://github.com/IBM/aihwkit}).  The toolkit is
  under active development and is centered around the concept of an
  ``analog tile'' which captures the computations performed on a
  crossbar array. Analog tiles are building blocks that can be used to extend existing network modules with analog components and compose arbitrary artificial neural networks (ANNs) using the flexibility of the \textsc{PyTorch} framework. Analog
  tiles can be conveniently configured to emulate a plethora of
  different analog hardware characteristics and their non-idealities, such
  as device-to-device and cycle-to-cycle variations, resistive device
  response curves, and weight and output noise. Additionally, the toolkit makes it possible to design custom unit cell configurations and to use advanced analog
  optimization algorithms such as Tiki-Taka. Moreover, the backward
  and update behavior can be set to ``ideal" to enable
  hardware-aware training features for chips that target inference
  acceleration only. To evaluate the inference accuracy of such chips
  over time, we provide statistical programming noise and drift models
  calibrated on phase-change memory hardware. Our new toolkit is
  fully GPU accelerated and can be used to conveniently estimate the
  impact of material properties and non-idealities of future
  analog technology on the accuracy for arbitrary ANNs.
\end{abstract}

\section{Introduction}
To cope with the ever increasing demand for computing resources in
artificial intelligence applications, dedicated hardware accelerators
for artificial neural network (ANN) training and inference have been
proposed recently. One promising future technology is the use of memristive 
crossbar arrays for accelerating the ubiquitous matrix-vector multiply and 
rank-update operations in
ANNs by employing in-memory computation of matrices stored as analog
quantities in tunable resistive elements~\cite{burr2017neuromorphic,
  haensch2018next,jain2019neural, Y2020sebastianNatNano}. While considerable run-time and
performance gain over today's technology is projected for analog
hardware in principle~\cite{gokmen2016acceleration}, noisy and
non-linear device characteristics as well as design restrictions such as
stationary weight matrices, demand for a new breed of ANN topologies
and algorithmic advances adapted to analog technology to unleash its full potential. An ideal outcome of such
``analog AI'' algorithmic endeavour would be reminiscent of the
increase in popularity of convolutional networks once it became clear that
their compute is ideally suited to be performed on graphic processing
units (GPUs).

To be viable, these new analog algorithms and optimized
network architectures need to be tested in a simulator that takes
analog hardware constraints and realistic material properties into
account.  Moreover, these simulations need to be applied on realistically
sized deep neural networks (DNNs) and datasets to make informed decisions about hardware
design and the expected accuracy.
\begin{figure}
  \centering
  \includegraphics[width=\columnwidth]{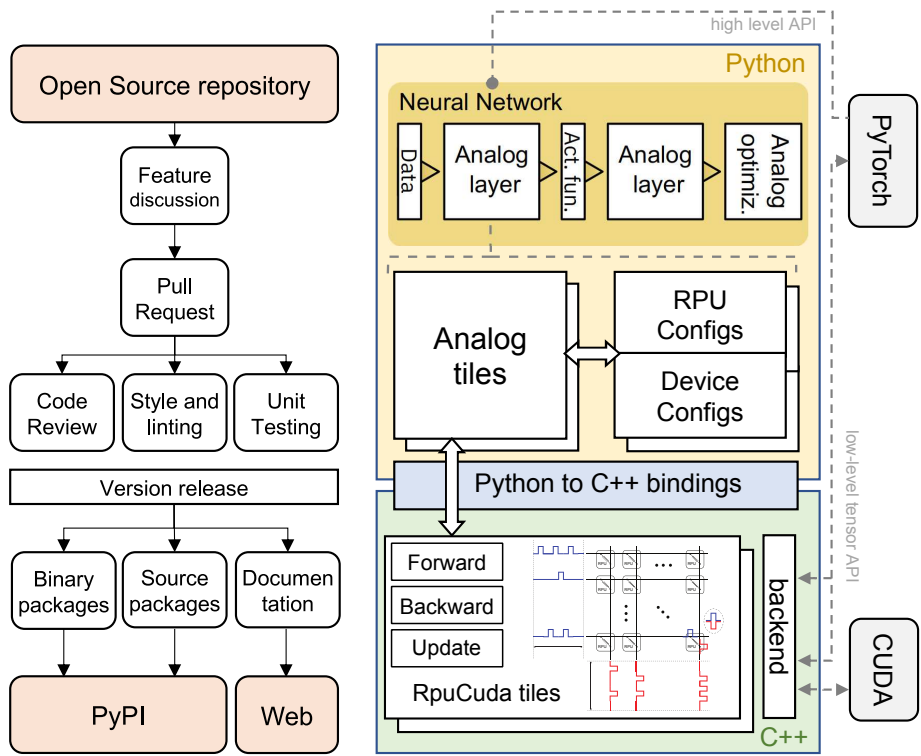}
  \caption{Repository and code structure of the IBM Analog Hardware Acceleration Kit.}
  \label{fig:code-structure}
\end{figure}

While there exists a number of current software packages that are
dedicated to simulating AI workloads on non-volatile memory elements,
notably \textsc{MLP-NeuroSim}~\cite{chen2017neurosim+} and
\textsc{RxNN} \cite{jain2020rxnn}, they are usually not tied to modern
AI frameworks, such as \textsc{PyTorch}~\cite{pytorch}. Such frameworks have become indispensable tools for AI researchers to build new
ANN architectures and implement AI workloads. For instance, while
\textsc{MLP-NeuroSim} \cite{chen2017neurosim+} has the ability to  handle a great number of
realistic hardware models, it is built around a small (by
today's standard) MNIST network example, hence lacking the tools to flexibly
implement new ANN topologies or simulate larger DNNs.  \textsc{RxNN} \cite{jain2020rxnn} is based on the rather outdated Caffe framework and only caters to analog chips dedicated to inference, lacking more
advanced algorithms and pulse update schemes that are needed for
training-enabled chip designs. A recent \textsc{PyTorch}
re-implementation of a subset of the \textsc{MLP-NeuroSim} package using
\textsc{PyTorch}, called \textsc{DNN+NeuroSim}~\cite{peng2020dnn+}, is
closest to our framework. However, it does not support fast custom
CUDA kernels for in-situ parallel analog update, and also lacks many of the aspects of modern code development, such as modularity, object oriented programming, proper Python binding
of fast C++ elements, unit testing, easy Python package installation
and so on. To address these limitations, we decided to open source IBM's internal
analog AI simulation core with a convenient \textsc{PyTorch} interface, hence providing adequate software tools for the advancement of algorithmic analog AI.

\section{Toolkit overview}
We provide a professionally maintained software kit that is easy to
install (as simple as \verb|pip install aihwkit|), has an extensive
documentation\footnote{\url{https://aihwkit.readthedocs.io/en/latest/}}
with a growing number of examples and tutorials, and has a well
integrated \textsc{PyTorch} experience providing a number of new
\verb|Analog| layers, such as linear (fully-connected) layer and convolutions (see~\figref{code-structure}). Under the hood, these layers perform the analog
hardware simulations using a highly optimized C++/CUDA core library
(\textsc{RpuCuda}) that is part of the package and made available under an Apache-2.0 Open Source license.  The core library is bound to the Python layer so that no
re-compilation is needed for defining models and parameters. Instead,
all C++ functions and analog hardware defining parameters are
conveniently accessible from Python. Figure~\ref{fig:code-structure} illustrates the architecture and code structure of the toolkit. 

Among others, we provide an \verb|AnalogLinear| layer, which
uses a single analog tile to compute the forward pass of a
fully-connected layer within a DNN. Defining a simple analog training simulations (with default device parameters) is similar to how \textsc{PyTorch} to define DNN layers (see example in \figref{example-listing}).

\begin{figure}
  \centering
\begin{minted}[framesep=6pt, frame=single, fontsize=\scriptsize]{python}
# Define crossbar (RPU) config
rpu_config = SingleRPUConfig(device=ReRamESPresetDevice())
# Define a single-layer network
model = AnalogLinear(4, 2, bias=True, rpu_config=config)
# Define analog-aware optimizer
opt = AnalogSGD(model.parameters(), lr=0.1)
opt.regroup_param_groups(model)
# Run the training
for epoch in range(100):
    pred = model(x) # forward pass
    loss = mse_loss(pred, y) 
    loss.backward() # backward pass
    opt.step()      # (analog pulsed) update
\end{minted}
\caption{Simple network definition and training with parallel pulsed update. Note that imports and data preparation are omitted for brevity (see example 01 in our GitHub repository).}
\label{fig:example-listing}
\end{figure}

\section{Analog tile model simulation }
The toolkit is centered around an ``analog tile'' that
corresponds to a 2D weight matrix that is stored on a non-volatile resistive
crossbar array and includes most pre- and post-processing (such as
dynamic input scaling and analog-digital converters) that are based on
(and extends) the resistive processing unit (RPU) definitions
in~\cite{gokmen2016acceleration}. The basic operation of this tile is a
matrix-vector product $  y_i = \sum_j w_{ij}x_j$, 
where the matrix elements $w_{ij}$ are assumed to be stored in the
crossbar (optionally together with the biases). The analog
matrix-vector product is, however, corrupted by noise and
non-idealities that are defined by the user through parameter
configurations of the tile during creation (called \verb|rpu_config|).

For instance, input/output/weight noise processes, analog-to-digital
conversion resolutions, or other non-idealities can be added so that
the effective computation simulated is e.g.\footnote{Consult the online
  documentation for a complete overview at \url{https://aihwkit.readthedocs.io}. } given by
\begin{equation}
  \label{eq:mat-vec-analog}
    y_i = f_\text{adc}\big(\sum_j(w_{ij} +
      \sigma_\text{w}\xi_{ij})(f_\text{dac}(x_j) 
      + \sigma_\text{inp}\xi_j)  + \sigma_\text{out}\xi_i\big)
\end{equation}
where $f_\text{adc}$ and $f_\text{dac}$ are discretization functions
(together with e.g. dynamic scaling and clipping) of the
digital-analog conversion process (as given by the hardware design),
and $\xi$ are Gaussian noise processes. In our tool, we generally opt
for a more abstract functional representation of the underlying hardware
(i.e. largely working with normalized units), since in this manner the
cause of accuracy loss of different hardware designs become
more apparent and specifications can be derived based on parameters
relevant to the DNN.

The RPU C++ library is the core of the high-performing implementation
of Eq.~\eqref{eq:mat-vec-analog}, which extensively uses dedicated and
fused GPU kernels. It provides the above basic model,
however, internal pre- and post processing (e.g. analog to digital
discretization) can be simply turned off or adapted by the user using
\textsc{PyTorch}'s functionality.

Within a DNN that is considered to be accelerated with analog crossbar
arrays, we generally assume that digital operations can be separated
from analog operations. For instance, by default, activation
functions are performed with digital computing using \textsc{PyTorch}'s 
library of floating point functions and assuming that analog result
values of a matrix vector product are digitized. Moreover, we currently
support only parallel read-out schemes and linear analog-digital
conversion. However, iterative read-out schemes and
shift-and-add for more complicated analog-digital converters could be
straightforwardly added in a similar manner using \textsc{PyTorch}.

While the forward pass includes cycle-to-cycle variations of the
resistive materials and circuit noise sources (see
Eq.~\ref{eq:mat-vec-analog}), for chips dedicated to ANN training,
backward and (potentially) update passes need to be similarly simulated
to be performed on analog as well. Thus, in our framework, backward
pass (the transposed version of the forward pass) can be configured to
contain noise and non-linearity similar to the forward pass. In
general, these parameters do not need to be the same, to reflect
hardware designs that differ in the noisiness in both directions.

A particular strength of our new software framework is the
highly-optimized pulse update with tune-able material pulse response
properties for chips that are dedicated to perform ANN training. For
that, we consider many algorithmic advances
\cite{gokmen2016acceleration, gokmen2020algorithm} and
use stochastic (or deterministic) pulse trains to incrementally perform
the theoretical rank update
\begin{equation}
  \label{eq:rank-update}
    w_{ij}^{(t+1)} = w_{ij}^{(t)} +  \lambda d_ix_j
\end{equation}
where $d_i$ are the back-propagated error signals. One possible way to
implement the update pass is to use stochastic pulse sequences of
variable or fixed length, where the probability of having a pulse is
proportional to the size of $x_j$ and $d_i$. In case of pulse
coincidences, the weight element at crosspoint $ij$ is updated by a
finite amount $\Delta w_{ij}$ (see \cite{gokmen2016acceleration}). 

Importantly, due to physical constraints, the amount $\Delta w_{ij}$
can vary in a systematic way from crosspoint to crosspoint
(device-to-device variations) or might have a systematic bias in one
or the other direction (up versus down pulses). Moreover, the step
$\Delta w_{ij}$ might depend in a non-linear way on the actual
conductance of the crosspoint value, depending on the physical
properties of the material used as resistive element.

\begin{figure}
  \centering
  \includegraphics[width=\columnwidth]{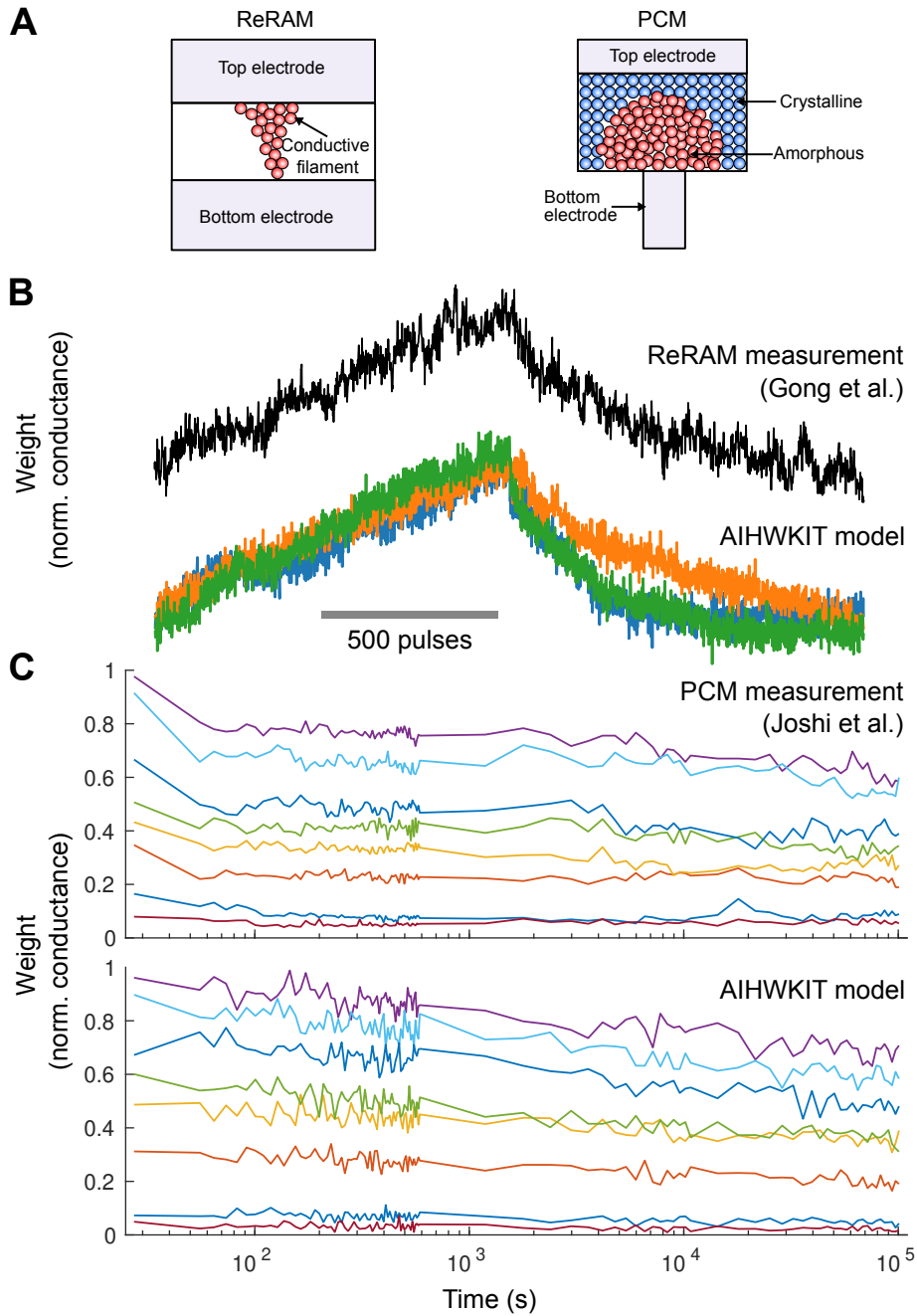}
  \caption{A) Schematic illustration of ReRAM and PCM devices. B) Pulse response of the experimental~\cite{gong2018signal} and simulated 
    ReRAM device (with device-to-device variations, write noise and, cycle-to-cycle variations).
    C) Experimentally obtained temporal evolution of PCM conductance~\cite{joshi2020accurate} compared to that simulated by the statistical noise model of our inference tile. Note that we assume currently that all weights are programmed at the same time in the simulation, whereas in the experiment devices converge at different iterations of programming and there will be jitter in times at which drift starts in individual devices. This effect leads to the slight difference noticeable for small times (less than 1000 seconds) between model and experiment.}
  \label{fig:reram}
  \label{fig:inference-noise}
\end{figure}

We thus provide a flexible way to choose the correct
update response and fit the update behavior to the material properties
under investigation, such as the amount of device-to-device variations
in various parameters and additional pulse-to-pulse variations.  We
also provide a number of presets calibrated on hardware data (e.g. for
ReRAM~\cite{gong2018signal} characteristics, see \figref{reram}). Example
response curves are shown in \figref{reram}.

Note that our framework differs significantly from the update approach
implemented by \textsc{DNN+NeuroSim}. For instance,
\textsc{DNN+NeuroSim} uses native \textsc{PyTorch} operators to
compute the accumulated gradient, which in case of batch size
larger than one or generally for convolutions will accumulate $p$
outer products in digital, and then only discretize the accumulated
result to calculate the pulse number and add pulse variations. This
means that the outer-product is actually done in digital and not analog.  This will result in a dramatic under-estimation of
the update noise properties for a system that performs the rank-1
update in place in parallel in analog~(as suggested by
\cite{gokmen2016acceleration}). Thus, we re-implemented
convolution and linear operators in our C++ core to ensure that
gradient accumulation happens using parallel update in analog memory even for
batch sizes larger than 1 or in case of convolutions.

Moreover, our framework draws the stochastic pulse trains and
applies the update pulses (in case of a coincidence) sequentially on
the devices, which are then updated according to the device model pulse
response curve. These pulsed updates are highly optimized 
for GPU compute, where we provide hundreds of custom CUDA
kernel templates that are automatically tuned for each tile, thus
selecting the fastest template on the fly. Consequently, the full
analog training simulation often takes only 2-5x longer than the
conventional floating-point training\footnote{Depending on DNN and device setting, e.g. 60s per epoch for VGG-8/CIFAR10 with parallel pulsed update versus 15s with floating point on a V100 GPU}. 

\section{Advanced device models and optimizers}
All the tile parameters are chosen by setting the resistive device
configuration. The default is having a single device per crosspoint. However, in many hardware designs, it might be advantageous to have multiple conductive elements within a ``unit cell''.  Therefore, we provide a mechanism to define unit cells having an arbitrary number
of devices (each potentially having different material properties),
which can be updated one-by-one or all together. This allows defining two uni-directional devices per cell, a multi-device cell or a
bi-directional single device cell with reference devices, by configuring the parameters of our analog tile accordingly.

Moreover, some analog optimization schemes have been proposed, where each analog tile of SGD is replaced with two analog tiles to create a coupled system by exchanging information between the two~\cite{gokmen2020algorithm} tiles. We
provide such constructs that can be
configured to test different and new analog inspired
optimization algorithms (see \figref{tiki-taka}).

Lastly, a number of temporal noise processes are available, such as
conductance decay, reset, and diffusion, which are all implemented
with systematic device-to-device variability. These temporal processes
are applied once per mini-batch and are thus assumed to be negligibly
small within the physical time a one mini-batch computation takes.

\section{Training for analog inference chips}
It has become increasingly apparent that accuracy on chips that are
designed for inference benefit greatly from ``hardware-aware''
training, where the DNN workload is trained (or fine-tuned) in software 
with some analog noise and non-idealities in the forward pass, but
with otherwise perfect backward and update pass~\cite{gokmen2019marriage,
  tsai2019inference, jain2019neural, joshi2020accurate}. Our framework
supports hardware-aware training for inference. In
particular, backward and update of a tile can simply be configured as
being ``perfect''. Moreover, we provide
weight noise functions for inference training, that
reversibly add noise onto weights during a mini-batch (in forward and
backward, but not during update pass) to train for noise resiliency.

Finally, we also provide a specific inference ``tile'' setting, which
supports adding carefully calibrated, conductance-dependent
programming noise, weight read noise and conductance drift onto a
trained networks' analog weights during inference time and provides
automatic global drift compensation~(see eg. \cite{joshi2020accurate}). The noise strength and temporal profile is calibrated on a 1M phase-change memory (PCM) device array \cite{joshi2020accurate} (see \figref{inference-noise}).

Thus, any network can be tested for realistic inference accuracy and
how the accuracy degrades over time.

\begin{figure}
  \centering
\begin{minted}[framesep=6pt, frame=single, fontsize=\scriptsize]{python}
# Define more complicated crossbar (RPU) config
rpu_config = UnitCellRPUConfig(
    device=TransferCompound(
        # Devices that compose the Tiki-taka compound.
        unit_cell_devices=[
            ReRamSBPresetDevice(dw_min_dtod=0.1),
            ReramSBPresetDevice(dw_min_std=0.2)],
        # Some adjustments to how to perform Tiki-Taka
        units_in_mbatch=True,   
        transfer_every=2))
\end{minted}
\caption{A more complex device configuration. This example implements the Tiki-taka modified SGD rule\cite{gokmen2020algorithm} for analog training. Once the \texttt{rpu\_config} is
  defined, the DNN training is identical to
  \figref{example-listing}. }
\label{fig:tiki-taka}

\end{figure}

\section{Extendability and open source contributions}
The reason of using \textsc{PyTorch} as an AI framework is to have a system to
flexibility define arbitrary DNN architectures. Our simulator
inherits this flexibility in defining networks with both analog and
digital components, where for the latter all features of \textsc{PyTorch} can be used. Moreover, because of the modular structure of our code
base, it is easy to extend and incorporate new features. For instance,
the inference noise models use Python classes and could be easily 
derived to implement other drift characteristics.

Device update behavior, on the other hand, is internally computed in
C++. This core C++ \textsc{RpuCuda} library is designed for speed but retains a modular structure to make it also extendable. While a number of abstract device models are
already provided, including a number of parameter presets,  new device models can be implemented relatively easily by inheriting most properties of existing
devices. In particular, if only the update response curve of a new
material needs to be changed, many optimized kernel routines can
be re-used, as only the update function itself needs to be written and
added to the code (in C++ for CPU and GPU).

In general, we follow an open source development model using GitHub standards, where we encourage contributions and discussions via an issue tracker, while using modern industry best-practices for software development (continuous integration, automated unit testing, code reviews) and packaging (semantic versioning, documentation, PEP adherence).

\section{Conclusion}
We present a new flexible open source software tool for simulating the
functional aspects of analog AI with a clear features roadmap. It is solely focused on the algorithmic development and functional verification for ANN training and inference on emerging analog chips. This toolkit is \emph{not} intended to estimate run time,
latency, or power performance of a hardware chip. While this is very important in its own right, performance aspects are (largely) irrelevant to the functional characterization and need dedicated and complimentary tools. 
We hope that this toolkit will make it easier for researchers
from AI hardware development and deep learning
communities, to develop new algorithms and design new ANN topologies
that are optimized for analog AI, hence experiencing the
exciting new area of analog AI.

\bibliographystyle{abbrv}
\bibliography{aicas2021_aihwkit_arxiv}

\begin{thebibliography}{10}

\bibitem{burr2017neuromorphic}
G.~W. Burr et~al.
\newblock Neuromorphic computing using non-volatile memory.
\newblock {\em Advances in Physics: X}, 2(1):89--124, 2017.

\bibitem{chen2017neurosim+}
P.-Y. Chen et~al.
\newblock {NeuroSim+}: An integrated device-to-algorithm framework for
  benchmarking synaptic devices and array architectures.
\newblock In {\em 2017 IEEE International Electron Devices Meeting (IEDM)},
  pages 6--1, 2017.

\bibitem{gokmen2020algorithm}
T.~Gokmen and W.~Haensch.
\newblock Algorithm for training neural networks on resistive device arrays.
\newblock {\em Frontiers in Neuroscience}, 14, 2020.

\bibitem{gokmen2019marriage}
T.~Gokmen, M.~J. Rasch, and W.~Haensch.
\newblock The marriage of training and inference for scaled deep learning
  analog hardware.
\newblock In {\em 2019 IEEE International Electron Devices Meeting (IEDM)},
  pages 22--3. IEEE, 2019.

\bibitem{gokmen2016acceleration}
T.~Gokmen and Y.~Vlasov.
\newblock Acceleration of deep neural network training with resistive
  cross-point devices: Design considerations.
\newblock {\em Frontiers in neuroscience}, 10:333, 2016.

\bibitem{gong2018signal}
N.~Gong and other.
\newblock Signal and noise extraction from analog memory elements for
  neuromorphic computing.
\newblock {\em Nature communications}, 9(2102), 2018.

\bibitem{haensch2018next}
W.~Haensch, T.~Gokmen, and R.~Puri.
\newblock The next generation of deep learning hardware: Analog computing.
\newblock {\em Proceedings of the IEEE}, 107(1):108--122, 2018.

\bibitem{jain2019neural}
S.~Jain et~al.
\newblock Neural network accelerator design with resistive crossbars:
  Opportunities and challenges.
\newblock {\em IBM Journal of Research and Development}, 63(6):10--1, 2019.

\bibitem{jain2020rxnn}
S.~Jain, A.~Sengupta, K.~Roy, and A.~Raghunathan.
\newblock {RxNN}: A framework for evaluating deep neural networks on resistive
  crossbars.
\newblock {\em IEEE Transactions on Computer-Aided Design of Integrated
  Circuits and Systems}, 2020.

\bibitem{joshi2020accurate}
V.~Joshi et~al.
\newblock Accurate deep neural network inference using computational
  phase-change memory.
\newblock {\em Nature Communications}, 11(2473), 2020.

\bibitem{pytorch}
A.~Paszke et~al.
\newblock Pytorch: An imperative style, high-performance deep learning library.

\bibitem{peng2020dnn+}
X.~Peng, S.~Huang, H.~Jiang, A.~Lu, and S.~Yu.
\newblock {DNN+NeuroSim V2.0}: An end-to-end benchmarking framework for
  compute-in-memory accelerators for on-chip training.
\newblock {\em arXiv:2003.06471}, 2020.

\bibitem{Y2020sebastianNatNano}
A.~Sebastian, M.~Le~Gallo, R.~Khaddam-Aljameh, and E.~Eleftheriou.
\newblock Memory devices and applications for in-memory computing.
\newblock {\em Nature Nanotechnology}, 15:529--544, 2020.

\bibitem{tsai2019inference}
H.~Tsai et~al.
\newblock Inference of long-short term memory networks at software-equivalent
  accuracy using 2.5 {M} analog phase change memory devices.
\newblock In {\em 2019 Symposium on VLSI Technology}, pages T82--T83. IEEE,
  2019.

\end{thebibliography}

\end{document}